\documentclass[runningheads]{llncs}
\usepackage[T1]{fontenc}
\usepackage{marvosym}
\usepackage{lipsum}
\usepackage{amsfonts}
\usepackage{amsmath}
\usepackage{tikz}
\usepackage{booktabs}
\usepackage{caption} 
\usepackage{hyperref}

\usepackage{graphicx}
\begin{document}
\title{Benchmark-Ready 3D Anatomical Shape Classification}
%
%
\author{Tomáš Krsička~\inst{(\textrm{\Letter})} \and
Tibor Kubík}
\hyphenation{proximity}

\authorrunning{T. Krsička and T. Kubík}
%
\institute{Brno University of Technology, Brno, Czech Republic\\
\email{xkrsic00@stud.fit.vut.cz}}
\maketitle              
\begin{abstract}
Progress in anatomical 3D shape classification is limited by the complexity of mesh data and the lack of standardized benchmarks, highlighting the need for robust learning methods and reproducible evaluation. We introduce two key steps toward clinically and benchmark-ready anatomical shape classification via self-supervised graph autoencoding. We propose \emph{Precomputed Structural Pooling} (\emph{PSPooling}), a non-learnable mesh pooling operator designed for efficient and structure-preserving graph coarsening in 3D anatomical shape analysis. PSPooling precomputes node correspondence sets based on geometric proximity, enabling parallelizable and reversible pooling and unpooling operations with guaranteed support structure. This design avoids the sparsity and reconstruction issues of selection-based methods and the sequential overhead of edge contraction approaches, making it particularly suitable for high-resolution medical meshes. To demonstrate its effectiveness, we integrate PSPooling into a self-supervised graph autoencoder that learns anatomy-aware representations from unlabeled surface meshes. We evaluate the downstream benefits on \emph{MedShapeNet19}, a new curated benchmark dataset we derive from MedShapeNet, consisting of 19 anatomical classes with standardized training, validation, and test splits. Experiments show that PSPooling significantly improves reconstruction fidelity and classification accuracy in low-label regimes, establishing a strong baseline for medical 3D shape learning. We hope that MedShapeNet19 will serve as a widely adopted benchmark for anatomical shape classification and further research in medical 3D shape analysis.
Access the complete codebase, model weights, and dataset information here:~\url{https://github.com/TomasKrsicka/MedShapeNet19-PSPooling}.
\end{abstract}
\section{Introduction}
Benchmark datasets have played a pivotal role in the development of computer vision. 
In 2D tasks, datasets such as MNIST~\cite{Lecun1998Mnist}, CIFAR-10~\cite{Krizhevsky2012Cifar}, and ImageNet~\cite{Deng2009ImageNet} provide a standard for reproducible and fair evaluation. 
Similarly, in 3D shape analysis, datasets such as the Princeton Shape Benchmark~\cite{SHilane2004Princeton}, ModelNet~\cite{Zhirong2015ModelNet}, and ShapeNet~\cite{Chang2015shapenet} provide standard benchmarks with increasing complexity and task diversity. While these datasets have enabled progress in shape classification, segmentation, and retrieval, they primarily consist of man-made CAD models with clean geometry. As such, they do not reflect the complexity of anatomical shapes encountered in medical imaging, where 3D surfaces reconstructed from CT or MRI scans often exhibit less regular topology, partial coverage, and high inter-subject variability. Medical datasets also face challenges related to privacy, labeling inconsistency, and data fragmentation~\cite{Arasteh2023Privacy,KARIMI2020noisyLabels}. Although MedShapeNet~\cite{Li2024medshapenet} was introduced to address these limitations, it remains task-agnostic and lacks predefined splits, limiting its immediate utility for specific standardized benchmark-ready tasks like medical shape classification.

Shape classification is a fundamental task in 3D geometry learning, with early approaches leveraging hand-crafted descriptors such as spin images~\cite{Johnson1999Spin}, shape histograms~\cite{Osada2002ShapeDistributions}, or spectral signatures~\cite{Reuter2009SpectralShape}, followed by traditional machine learning classifiers.  Volumetric CNNs and point-based networks such as PointNet~\cite{Charles2017PointNet} and 3dShapeNets~\cite{Zhirong2015ModelNet} allowed direct input of 3D shapes, but often suffered from inefficiencies due to data sparsity or loss of topological structure. More recently, graph neural networks (GNNs)~\cite{Scarselli2029gnn} have emerged as a natural framework to represent and learn from 3D meshes where vertices and edges encode geometric and structural relationships. Variants such as graph convolutional networks (GCNs)~\cite{kipf2017gcn} and graph attention networks (GATs)~\cite{Velickovic2017GAT} allow flexible aggregation over irregular domains, and pooling schemes based on edge contraction~\cite{diehl2019EdgePooling}, node selection~\cite{Gao2019GraphUnets,Lee2019SAGPooling}, or hierarchical clustering~\cite{Duval2022NodeClustering} have been explored to handle variable mesh resolution. 

To improve generalization in label-scarce settings, self-supervised pretraining can be leveraged, provided that effective pooling and unpooling operations are available to capture and reconstruct meaningful structure~\cite{Charte_2020}.
Local pooling methods aim to reduce graph resolution by aggregating or selecting subsets of nodes.\emph{ Selection-based operators} such as gPool~\cite{Gao2019GraphUnets} and SAGPool~\cite{Lee2019SAGPooling} score and retain top-ranked nodes, discarding the rest. Although computationally efficient and parallelizable, they often result in sparsely connected subgraphs because high-scoring nodes may not share edges in the original graph. This limits the message passing capacity in deeper GNN layers. Their corresponding unpooling strategies typically rely on restoring features to their original positions while filling in discarded nodes with placeholders, which fails to reconstruct the local structure and leads to information loss.
\emph{Edge contraction methods} by contrast, operate by merging node pairs along high-scoring edges, preserving local connectivity and typically producing better structural integrity in the downsampled graph. For instance, methods like EdgePool~\cite{diehl2019EdgePooling} score edges using node feature combinations and iteratively contract non-overlapping edges. This preserves more of the original graph’s topology and maintains denser connections than selection pooling, which is beneficial for maintaining global information flow. However, these methods suffer from serious computational limitations. Because contraction decisions are made sequentially and must respect merge conflicts, the algorithm is inherently nonparallelizable. This renders the approach impractically slow for large graphs like 3D medical shapes.

\pagebreak
We claim the contributions of this paper to be twofold:
\begin{itemize}
    \item First, we introduce \emph{MedShapeNet19}, a curated benchmark dataset for medical shape classification derived from the publicly available MedShapeNet~\cite{Li2024medshapenet}. It retains 19 anatomical classes from the original collection, each represented by 800 surface mesh samples reconstructed from clinical imaging data. By focusing on a well-balanced subset with predefined splits, MedShapeNet19 offers a realistic yet manageable benchmark for anatomical shape classification, capturing the geometric complexity and variability often encountered in medical practice.
    \item Second, we introduce \emph{Precomputed Structural Pooling} (\emph{PSPooling}), a mesh pooling method that preserves the local anatomical structure by incorporating geometric context into the coarsening process. To improve performance in limited-label and heterogeneous-data scenarios common in medical imaging, we leverage self-supervised pretraining using a graph autoencoder to enhance downstream classification, particularly when labeled anatomical data are scarce.
    Based on this approach, we train a model as the baseline standard method for the MedShapeNet19 benchmark, establishing a consistent foundation for future comparisons.
\end{itemize}

\section{MedShapeNet19: A Curated Benchmark Set for Medical Shape Classification}
To address the limitations of existing 3D medical shape datasets, we present \textit{MedShapeNet19}, a curated and balanced subset of the original MedShapeNet~\cite{Li2024medshapenet}, tailored for anatomical classification tasks. While MedShapeNet itself contains over 100{,}000 samples, their distribution is not uniform. For instance, more than 40{,}000 samples are allocated to various \textit{rib} or \textit{vertebrae} classes. Furthermore, the dataset includes left and right variants for each applicable anatomical structure with some instances being represented in multiple classes. Due to the uneven sample distribution in the original dataset, a class selection process was performed to ensure balance and reduce redundancy. Classes corresponding to the same general anatomical structure and left-right variations were merged. Classes with extreme face and vertex counts (e.g., \textit{skull}, \textit{brain}), or those missing complete organ representations (e.g., \textit{heart}, \textit{lungs}), were excluded. From the remaining classes, we considered only those that had more than 800 valid samples. Classes of the \textit{blood vessel} category were omitted due to insufficient representation.

Invalid samples were filtered through a two-step process. Extremely small files (on the order of hundreds of bytes) were automatically discarded, while larger files (hundreds of kilobytes) were further analyzed. Empirical criteria showed that meshes with fewer than 1{,}000 faces contained useful data only if they formed valid manifolds. Non-manifold samples with higher face counts were retained because of their structural relevance.

\begin{figure}[h!]
\centering
\includegraphics[width=\textwidth]{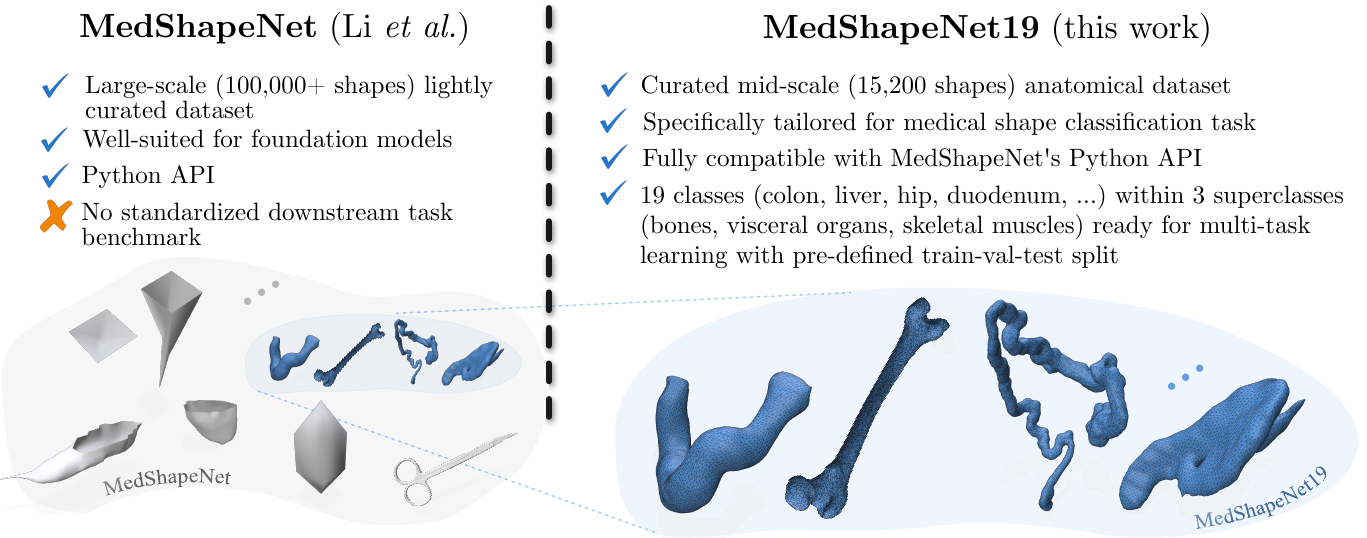}
\caption{\textbf{MedShapeNet19: a curated benchmark dataset for 3D medical shape classification derived from MedShapeNet~\cite{Li2024medshapenet}.}
We refine an initially large-scale and weakly curated dataset by filtering erroneous images, removing non-body structures such as instruments, and merging semantically similar labels. 
}
\label{fig:medshapenet19-teaser}
\end{figure}

As shown in Figure~\ref{fig:medshapenet19-teaser}, the final \textit{MedShapeNet19} dataset consists of 15{,}200 3D anatomical models, spanning 19 classes grouped into three superclasses: bones, visceral organs, and skeletal muscles. Each class contains 800 samples. For our experiments, to standardize geometric complexity across samples, we applied quadratic mesh decimation, reducing the number of faces to approximately 10{,}000 per model.

\section{Boosting Medical Shape Classification Performance via Graph Autoencoders with PSPooling}
Building upon the MedShapeNet19 benchmark introduced in the previous section, we now describe our approach for improving anatomical shape classification using graph autoencoders with PSPooling. Each shape from the benchmark is represented as an undirected graph and annotated with a semantic class label. 
The goal is to learn a model that maps each input graph to its corresponding class label. We employ a graph neural network (GNN) encoder $f_\theta$ that computes a latent representation of the mesh, followed by a classification head that predicts the label based on this representation. This setup enables mesh-level semantic classification trained using supervised learning on labeled data.

To enhance the model's ability to generalize in label-scarce scenarios, we first perform self-supervised pretraining. Specifically, we adopt an encoder-decoder architecture in which the encoder $f_\theta$ produces a latent embedding of the mesh, and a decoder $g_\phi$ reconstructs structural or geometric aspects of the original input. This pretraining phase allows the encoder to capture generic anatomy-aware mesh features without relying on labels.
When applied to graphs, this geometric prior can improve downstream classification by enforcing structure-aware feature learning. However, it necessitates effective pooling and unpooling operators. Traditional unpooling methods like SAGPool~\cite{Lee2019SAGPooling} may fail to recover structural information lost during pooling, limiting reconstruction fidelity, and, by extension, the benefit of pretraining.

\begin{figure}[h!]
\centering
\includegraphics[width=\textwidth]{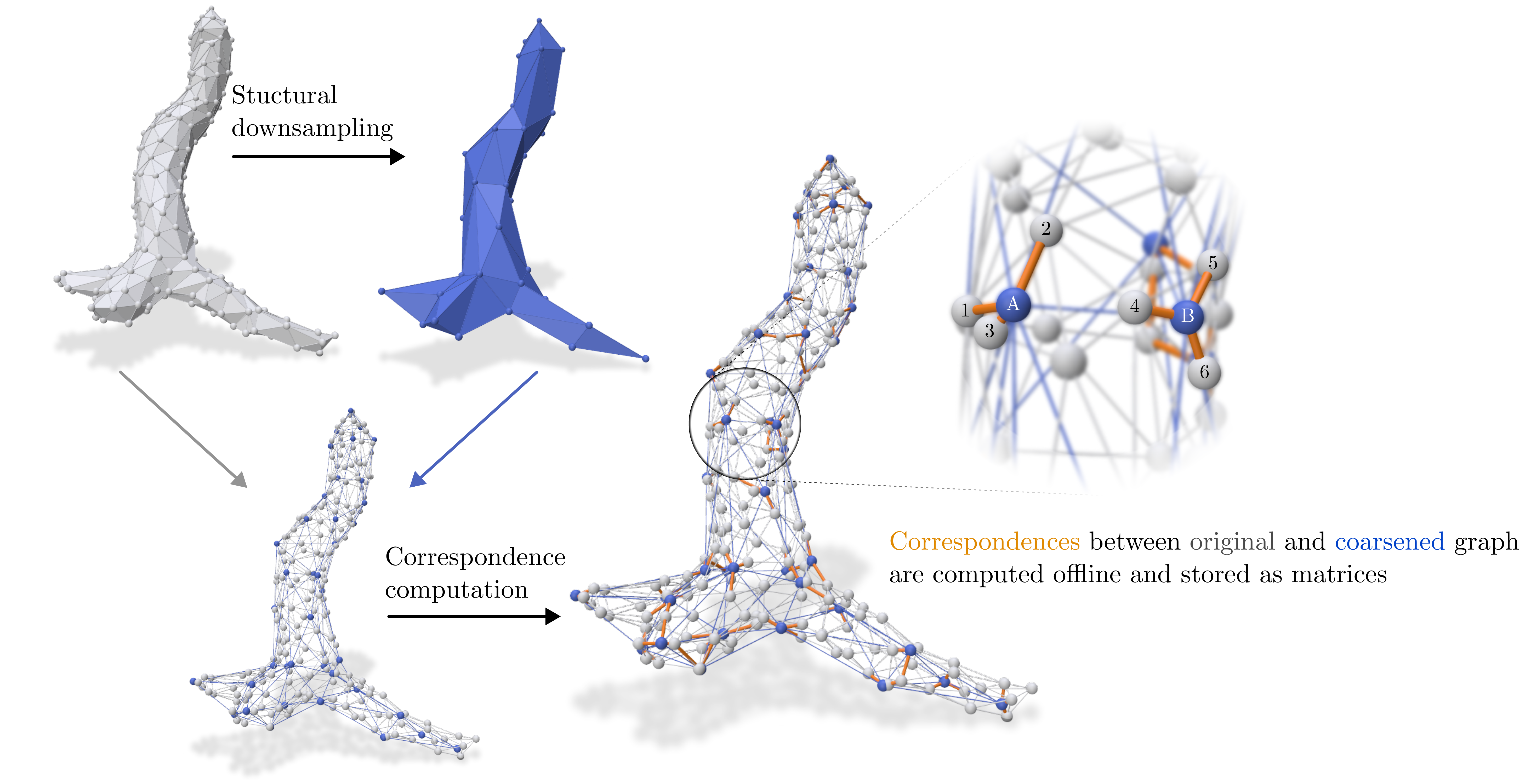}
\caption{\textbf{Method outline of our proposed Precomputed Structural Pooling operator.}
Node correspondences are precomputed between successive levels of the subsampled meshes. Node A gathers features from nodes 1,\,2 and 3. Node~B from nodes 4,\,5 and 6.
}
\label{fig:PsPooling-teaser}
\end{figure}

To that extent, we propose \emph{Precomputed Structural Pooling} (\emph{PSPooling}), a non-learnable operator tailored for graph domains that support meaningful structural downsampling (e.g., 3D meshes). Unlike trainable pooling methods, PSPooling precomputes the graph coarsening and correspondence mappings prior to training, enabling fully parallelizable pooling and unpooling operations that maintain graph connectivity throughout. This process is illustrated in Figure~\ref{fig:PsPooling-teaser}.

Given an input graph $G = (V, E)$ with node features $X \in \mathbb{R}^{|V| \times d}$, we define a coarsened version $G' = (V', E')$ with a reduced node set and features $X' \in \mathbb{R}^{|V'| \times d}$. Each coarse node $v'_i \in V'$ is associated with a fixed-size subset of fine nodes $v_j \in V$ via a correspondence function
\begin{equation}
    \mathcal{R} : V' \to 2^V.
\end{equation}
This relation defines the local support from which the features in $X$ are aggregated into $X'$. For computational efficiency and regularity, the cardinality of each correspondence set is bounded: \(|\mathcal{R}(v'_i)| \leq k_S\), where $k_S$ is a predefined hyperparameter (e.g., the $k$ nearest neighbors in geodesic space).

\noindent The pooling operation is implemented as a sparse matrix multiplication using a precomputed weight matrix $P \in \mathbb{R}^{|V'| \times |V|}$, with entries
\begin{equation}
    P_{ij} = 
    \begin{cases}
    w_{ij}, & \text{if } v_j \in \mathcal{R}(v'_i), \\
    0, & \text{otherwise},
    \end{cases}
\end{equation}
where weights $w_{ij}$ can reflect domain-specific heuristics such as geodesic distance. To preserve feature scale, the rows of $P$ are normalized to sum to one:
\begin{equation}
    P_{ij} \leftarrow \frac{P_{ij}}{\sum_{k \in \mathcal{R}(v'_i)} P_{ik}}.
\end{equation}
The pooled features are then computed as
\begin{equation}
    X' = P X.
\end{equation}

\noindent In settings where structural invariance is desired, PSPooling also supports a non-linear variant via element-wise max pooling:
\begin{equation}
    X'_i = \max_{j \in \mathcal{R}(v'_i)} X_j.
\end{equation}
\subsubsection{Unpooling and Support Balancing}
The unpooling operator redistributes the features from $V'$ back to $V$ using the transpose of the unnormalized pooling matrix $P^\top$, which is then row normalized. Due to the inherent difference in the number of nodes between successive levels of mesh subsampling, the number of pooling correspondences typically exceeds that of unpooling correspondences. To ensure adequate representation of fine-level nodes during unpooling, the size of the initial correspondence set $\mathcal{R}(v'_i)$ can initially be increased during precomputation. The transposed relation $\mathcal{R}^\top$ is then derived from this augmented set, and both $\mathcal{R}$ and $\mathcal{R}^\top$ are subsequently truncated to retain at most $k_S$ connections per node. This truncation balances the support between the pooling and unpooling operations.

\subsection{Model Overview}

Each model follows a shared architectural pattern, consisting of an initial stack of GAT layers to project the 6-dimensional input node features into a higher-dimensional latent space. These layers are followed by a series of pooling and GAT blocks, each reducing the spatial resolution while increasing feature dimensionality. After the final pooling stage, a global attention-based aggregation module produces a graph-level representation that is used for downstream classification or decoding.

To support reconstruction, a 2-dimensional auxiliary feature vector is extracted from each node immediately prior to global pooling. These narrow per-node embeddings are stored and later concatenated to the reconstructed node features during the first stage of unpooling. This mechanism helps reintroduce the spatial structure lost during the pooling and improves the effectiveness of fixed unpooling operations.

The decoder mirrors the encoder architecture, applying the corresponding unpooling operator and GAT layers at each stage. A four-layer MLP is used to map the final decoded features back to the vertex coordinates. All other architectural components are held fixed across all pooling variants to ensure a fair comparison.

Three architecture sizes, namely \texttt{S}, \texttt{M}, and \texttt{L}, are evaluated. They differ in network depth and bottleneck dimensionality: architectures \texttt{S} and \texttt{M} use a pooling depth of 2, with bottleneck sizes of 256 and 512 channels, respectively, while \texttt{L} uses a pooling depth of 3 and a 1024-dimensional bottleneck. The pooling and unpooling layers are implemented using either the proposed PSPooling operator or the alternative SAG pooling, with all other settings kept constant.

\section{Experiments and Results}
To ensure a controlled and reproducible evaluation, all experiments are conducted on the MedShapeNet19 dataset. We construct a standardized 8:1:1 split of the dataset into training, validation, and test partitions, respectively. 

To rigorously isolate the effect of the pooling strategy, all pooling operators are evaluated within identical architectural contexts. That is, we employ the same encoder-decoder configuration across experiments, varying only the pooling and corresponding unpooling components. This enables a direct comparison between the proposed PSPooling operator and existing alternatives such as SAG pooling~\cite{Lee2019SAGPooling}. 

All models are implemented in PyTorch 2.5.1 with PyTorch Geometric 2.6.1 and trained using CUDA 12.1 on GPU-equipped nodes allocated via metacentrum.cz. Each training run was assigned a minimum of 40\,GB of GPU memory, typically utilizing NVIDIA A40, A100, or H100 accelerators. Training management and logging were handled through PyTorch Lightning 2.5.0. All models were trained for a fixed number of epochs with early stopping based on validation loss.

\subsection{Autoencoder Mesh Reconstruction}

We first evaluate the ability of the autoencoder to reconstruct the 3D mesh structure from a compressed latent representation. This experiment highlights the strengths and weaknesses of each pooling strategy when used in a generative task.

\begin{figure}[h!]
\centering
\includegraphics[width=0.86\linewidth]{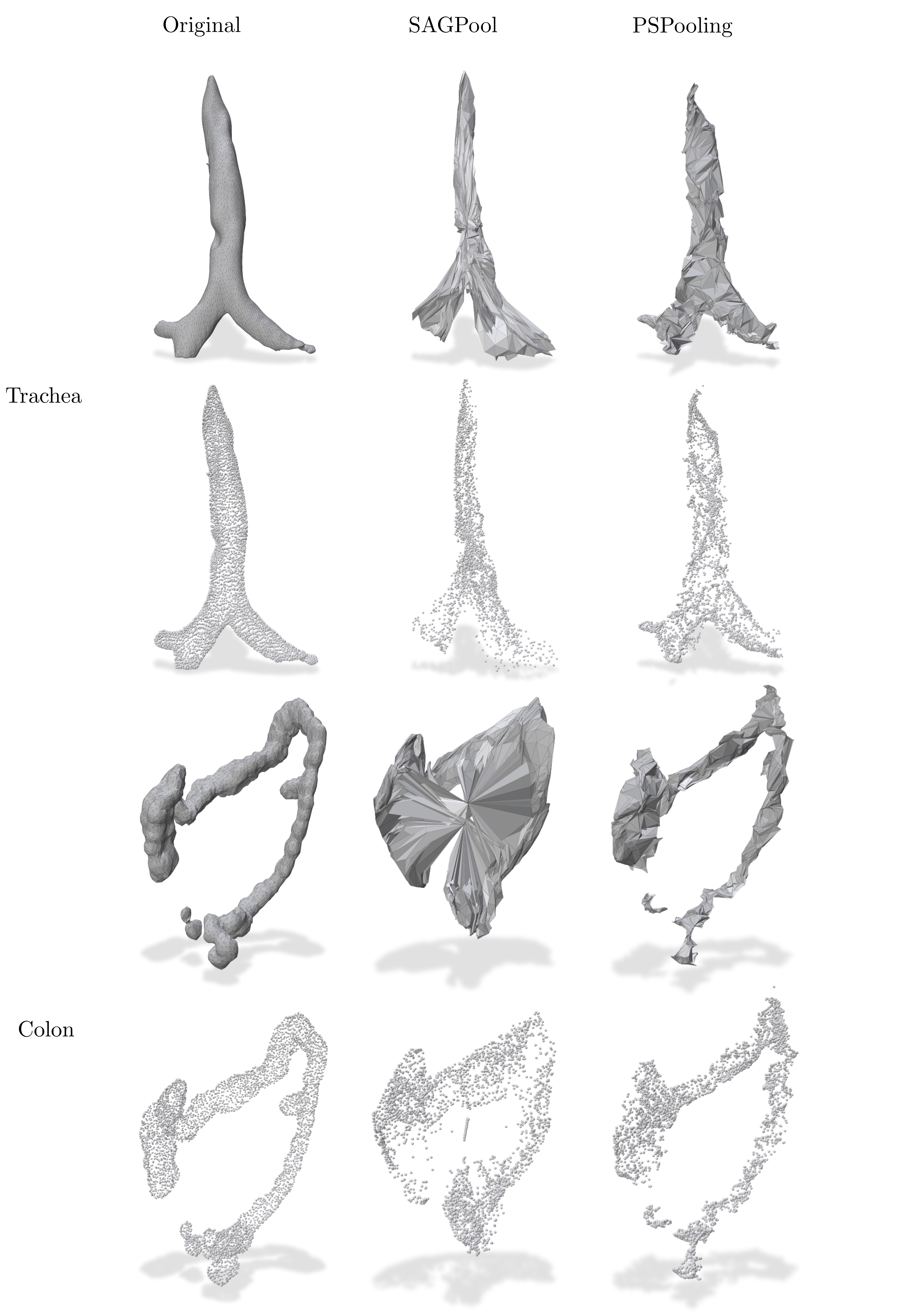}
\caption{\textbf{Comparison of reconstruction quality between architectures utilizing SAG pooling and PSPooling.} PSPooling reconstructions show better detail retention, preserve disconnected components and prevent formation of vertex clustering artifacts. 
}
\label{fig:reconstructions}
\end{figure}

Visual inspection reveals that PSPooling consistently results in smoother and more coherent surface reconstructions. It avoids the formation of dense vertex clusters, a frequent artifact observed in SAG pooling-based models. These clusters, especially when located away from the mesh body, generate large, flat, phantom faces due to fixed mesh connectivity and degenerate vertex positions. In contrast, PSPooling reconstructions maintain a more uniform spatial distribution of vertices with better preservation of small details, including disconnected components. Reconstruction examples are shown in Figure~\ref{fig:reconstructions}.

\subsection{Transfer Learning via Pretraining}

Next, we evaluate whether self-supervised pretraining using an autoencoder can improve performance on downstream classification tasks. To this end, encoders are first trained to reconstruct mesh data, then frozen and used to extract embeddings for a linear classification layer trained on labeled data.

\begin{figure}[h!]
\centering
    \begin{tikzpicture}
      \node[anchor=south west, inner sep=0] (image) at (0,0) {\includegraphics[width=\linewidth]{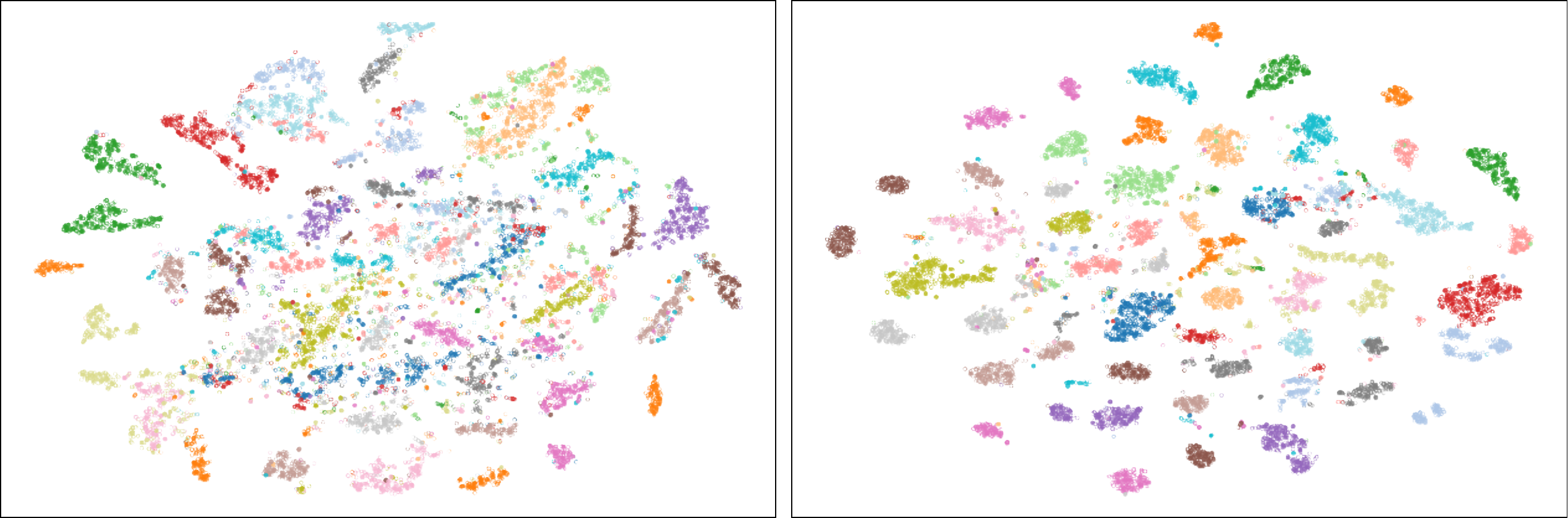}};
      \begin{scope}[x={(image.south east)}, y={(image.north west)}]
        \node[align=center] at (0.25, 1.05) {SAGPool};
        \node[align=center] at (0.75, 1.05) {PSPooling};
      \end{scope}
    \end{tikzpicture}
\caption{\textbf{t-SNE visualizations of latent spaces produced by autoencoders employing SAGPool (left) and PSPooling (right).} The latent space from a network that uses PSPooling exhibits more pronounced clustering tendencies.
}
\label{fig:tsne}
\end{figure}

Results presented in Table~\ref{tab:downstream_full} indicate that encoders pretrained with PSPooling yield classification accuracy comparable to fully supervised training, demonstrating the effectiveness of the learned representations. In contrast, encoders trained with SAG pooling show a clear drop in accuracy when frozen, particularly for larger architectures. t-SNE visualizations shown in Figure~\ref{fig:tsne} confirm that PSPooling yields more compact and well-separated class clusters, even without fine-tuning.

\begin{table}[ht]
    \centering
    \begin{tabular}{l@{\hskip 0.4in}c@{\hskip 0.2in}c@{\hskip 0.2in}c c}
        \toprule
        \textbf{Model Type} & \texttt{S} & \texttt{M} & \texttt{L} \\
        \midrule
        Supervised only on classification (SAGPool)      & 96.4\% & 97.0\% & 96.7\% \\
        Supervised only on classification (PSPooling)             & 96.5\% & \textbf{97.8}\% & 97.6\% \\
        Pretrained (SAGPool)  & 91.1\% & 92.6\% & 89.7\% \\
        Pretrained (PSPooling)         & 96.1\% & 96.0\% & \textbf{97.0}\% \\
        \bottomrule
    \end{tabular}
    
    \caption{\textbf{Classification accuracy across different architecture sizes and training strategies.} The classifiers trained over a~frozen encoder using the PSPooling operator perform on par with a~supervised classifier with the same pooling operator.  Highlighted are the best results for a direct classifier and a pretrained classifier.}
    \label{tab:downstream_full}
\end{table}

\subsection{Limited Supervision Scenario}

\begin{figure}[h!]
\centering
\includegraphics[width=\linewidth]{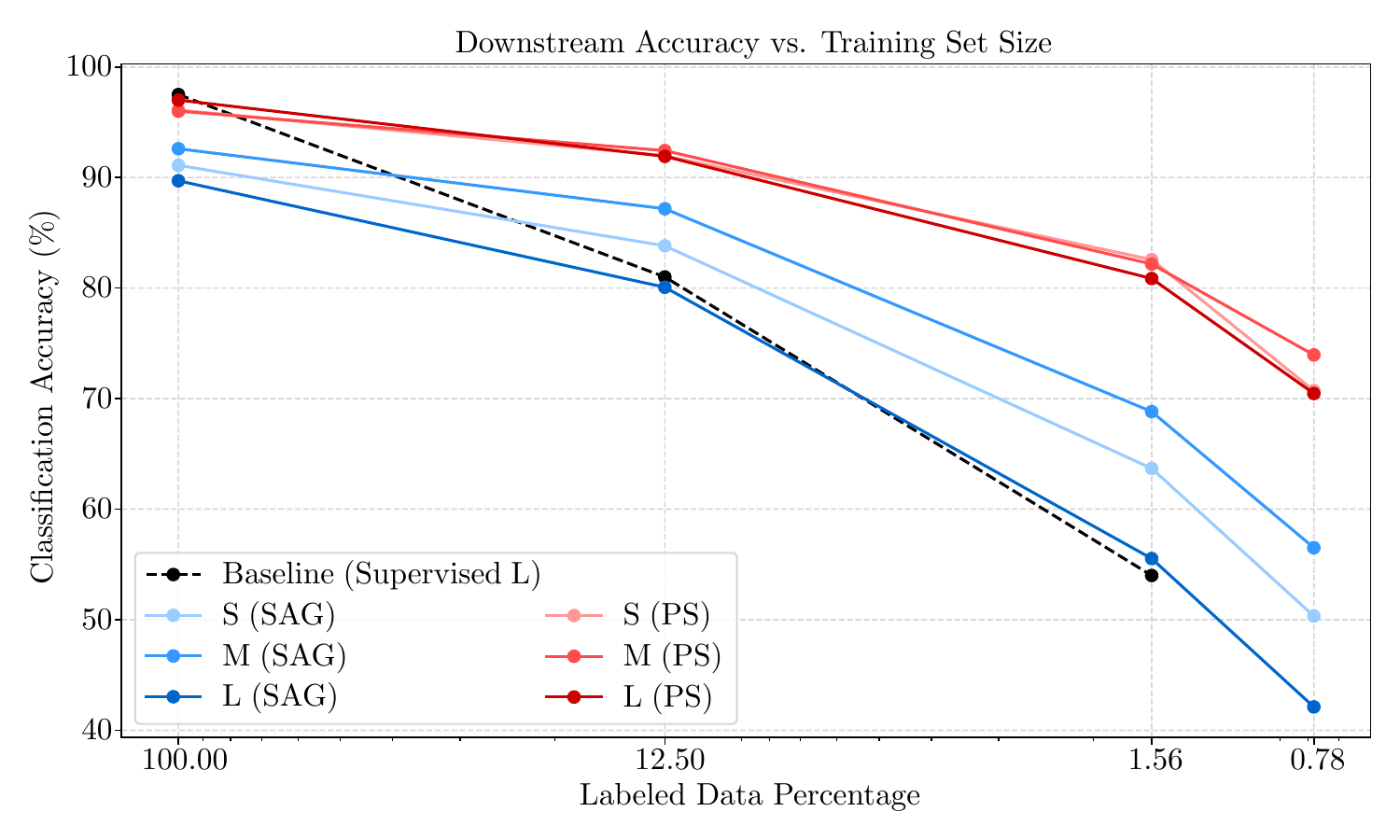}
\caption{\textbf{Downstream classification accuracy comparison of pooling strategies across various dataset fractions.} While both methods show performance degradation with reduced training data, networks using PSPooling (red curves) demonstrate greater robustness under limited supervision, showing a smaller performance drop compared to SAGPool (blue curves). 
}
\label{fig:reduced_classification}
\end{figure}

We further investigate classification under label-scarce conditions by pretraining and fine-tuning classifiers on progressively smaller subsets of the labeled training data (Figure~\ref{fig:reduced_classification}). This simulates practical scenarios such as medical datasets, where labeled samples are limited.

Under extreme scarcity (0.78\% of training data), PSPooling models retain more than 70\% classification accuracy, outperforming a fully supervised baseline trained on twice as much labeled data. Even at moderate label fractions (1.56\% or 12.5\%), PSPooling models remain competitive, while SAG pooling models suffer from sharp performance drops. This highlights the robustness and generalization capability of PSPooling-based encoders.

Interestingly, smaller PSPooling models outperform deeper ones in low-label settings, suggesting that deeper models may allocate capacity to finer details irrelevant for class discrimination. This trade-off is visible in both accuracy trends and feature-space visualizations.
\section{Conclusion}

In this paper, we introduce MedShapeNet19, a filtered and balanced subset of MedShapeNet, curated for standardized evaluation in 3D shape analysis. It comprises 15,200 samples across 19 anatomical classes. Building on this benchmark, we propose PSPooling, a fully precomputed pooling operator for mesh-based learning, designed for self-supervised graph pretraining. PSPooling reduces graph resolution using static, geodesic-based weights while preserving global shape structure. Experimental results show that: (1) PSPooling improves classification accuracy and structural sensitivity; (2) it enables cleaner and more faithful reconstructions than selection-based pooling; and (3) pretrained models using PSPooling achieve performance on par with or exceeding fully supervised models, especially in label-scarce scenarios, matching accuracy with up to eight times fewer labeled samples.

Future work could explore attention-based strategies and the incorporation of physical and topological priors during pretraining. While this work focused on classification, PSPooling could also benefit other downstream tasks such as segmentation or reconstruction, where preserving and recovering geometric structure through pooling and unpooling may improve spatial accuracy and representational quality.
Additionally, we hope that future research will leverage MedShapeNet19 as a standardized benchmark dataset for advancing medical shape classification, facilitating comparative evaluation of novel architectures and training paradigms.

\begin{credits}
\subsubsection{\ackname} Computational resources were provided by the e-INFRA CZ project (ID:90254), supported by the Ministry of Education, Youth and Sports of the Czech Republic.

\subsubsection{\discintname}
The authors have no competing interests to declare that are
relevant to the content of this article. 
\end{credits}
%
%
%
\bibliographystyle{splncs04}
\bibliography{mybib}

\begin{thebibliography}{10}
\providecommand{\url}[1]{\texttt{#1}}
\providecommand{\urlprefix}{URL }
\providecommand{\doi}[1]{https://doi.org/#1}

\bibitem{Chang2015shapenet}
Chang, A.X., Funkhouser, T., Guibas, L., Hanrahan, P., Huang, Q., Li, Z., Savarese, S., Savva, M., Song, S., Su, H., Xiao, J., Yi, L., Yu, F.: {ShapeNet: An Information-Rich 3D Model Repository}. ArXiv abs/1512.03012  (2015)

\bibitem{Charles2017PointNet}
Charles, R.Q., Su, H., Kaichun, M., Guibas, L.J.: Pointnet: Deep learning on point sets for 3d classification and segmentation. In: IEEE Conference on Computer Vision and Pattern Recognition (CVPR) (2017)

\bibitem{Charte_2020}
Charte, D., Charte, F., del Jesus, M.J., Herrera, F.: An analysis on the use of autoencoders for representation learning: Fundamentals, learning task case studies, explainability and challenges. Neurocomputing  (2020)

\bibitem{Deng2009ImageNet}
Deng, J., Dong, W., Socher, R., Li, L.J., Li, K., Fei-Fei, L.: Imagenet: A large-scale hierarchical image database. In: IEEE Conference on Computer Vision and Pattern Recognition (CVPR) (2009)

\bibitem{diehl2019EdgePooling}
Diehl, F., Brunner, T., Le, M.T., Knoll, A.: Towards graph pooling by edge contraction. In: ICML workshop on learning and reasoning with graph-structured data (2019)

\bibitem{Duval2022NodeClustering}
Duval, A., Malliaros, F.D.: Higher-order clustering and pooling for graph neural networks. ACM International Conference on Information \& Knowledge Management  (2022)

\bibitem{Gao2019GraphUnets}
Gao, H., Ji, S.: Graph u-nets. IEEE Transactions on Pattern Analysis and Machine Intelligence (PAMI)  (2019)

\bibitem{Johnson1999Spin}
Johnson, A., Hebert, M.: Using spin images for efficient object recognition in cluttered 3d scenes. IEEE Transactions on Pattern Analysis and Machine Intelligence (PAMI)  (1999)

\bibitem{KARIMI2020noisyLabels}
Karimi, D., Dou, H., Warfield, S.K., Gholipour, A.: Deep learning with noisy labels: Exploring techniques and remedies in medical image analysis. Medical Image Analysis  (2020)

\bibitem{kipf2017gcn}
Kipf, T., Welling, M.: Semi-supervised classification with graph convolutional networks. ArXiv abs/1609.02907  (2016)

\bibitem{Krizhevsky2012Cifar}
Krizhevsky, A.: Learning multiple layers of features from tiny images. University of Toronto  (2012)

\bibitem{Lecun1998Mnist}
Lecun, Y., Bottou, L., Bengio, Y., Haffner, P.: Gradient-based learning applied to document recognition. Proceedings of the IEEE  (1998)

\bibitem{Lee2019SAGPooling}
Lee, J., Lee, I., Kang, J.: Self-attention graph pooling. ArXiv abs/1904.08082  (2019)

\bibitem{Li2024medshapenet}
Li, J., Zhou, Z., Yang, J., Pepe, A., Gsaxner, C., Luijten, G., Qu, C., Zhang, T., Chen, X., Li, W., Wodzinski, M., Friedrich, P., Xie, K., Yuan, J., Ambigapathy, N., Nasca, E., Solak, N., Melito, G.M., Vu, V., Egger, J.: Medshapenet – a large-scale dataset of 3d medical shapes for computer vision. Biomedical Engineering / Biomedizinische Technik  (2024)

\bibitem{Osada2002ShapeDistributions}
Osada, R., Funkhouser, T., Chazelle, B., Dobkin, D.: Shape distributions. ACM Transactions on Graphics (TOG)  (2002)

\bibitem{Reuter2009SpectralShape}
Reuter, M., Biasotti, S., Giorgi, D., Patanè, G., Spagnuolo, M.: Discrete laplace–beltrami operators for shape analysis and segmentation. Computers \& Graphics  (2009)

\bibitem{Scarselli2029gnn}
Scarselli, F., Gori, M., Tsoi, A., Hagenbuchner, M., Monfardini, G.: The graph neural network model. IEEE transactions on neural networks / a publication of the IEEE Neural Networks Council  (2009)

\bibitem{SHilane2004Princeton}
Shilane, P., Min, P., Kazhdan, M., Funkhouser, T.: The princeton shape benchmark. In: Proceedings Shape Modeling Applications, 2004. (2004)

\bibitem{Arasteh2023Privacy}
Tayebi~Arasteh, S., Ziller, A., Kuhl, C., Makowski, M., Nebelung, S., Braren, R., Rueckert, D., Truhn, D., Kaissis, G.: Private, fair and accurate: Training large-scale, privacy-preserving ai models in medical imaging. Communications Medicine  (2024)

\bibitem{Velickovic2017GAT}
Velickovic, P., Cucurull, G., Casanova, A., Romero, A., Lio’, P., Bengio, Y.: Graph attention networks. ArXiv abs/1710.10903  (2017)

\bibitem{Zhirong2015ModelNet}
Wu, Z., Song, S., Khosla, A., Yu, F., Zhang, L., Tang, X., Xiao, J.: 3d shapenets: A deep representation for volumetric shapes. In: IEEE Conference on Computer Vision and Pattern Recognition (CVPR) (2015)

\end{thebibliography}
%
%
%
%
%
\end{document}